\documentclass[12pt]{article}%
\usepackage{amsmath}
\usepackage{amsfonts}
\usepackage{amssymb}
\usepackage{graphicx}
\usepackage{url,hyperref}
\providecommand{\U}[1]{\protect \rule{.1in}{.1in}}

\begin{document}

\title{Surv-IPTB: An Attention-Based Model for Estimating Individual Probability of Treatment
Benefit with Survival Data}
\author{Lev V. Utkin, Stanislav K. Kogan and Andrei V. Konstantinov\\ 
\small{Higher School of Artificial Intelligence Technologies} \\ \small{Peter the Great St.Petersburg Polytechnic University, St.Petersburg, Russia}
}
\date{}
\maketitle

\begin{abstract}
This work presents a novel attention-based framework for estimating the
Individual Probability of Treatment Benefit (IPTB) in survival analysis
contexts. The proposed model, called Surv-IPTB, directly quantifies the probability that a specific
patient will experience extended survival time under treatment versus control.
We reformulate IPTB estimation as a binary classification problem, leveraging
pairwise patient comparisons across treatment and control cohorts. The
framework incorporates a principled handling of right-censored observations
through imprecise probability representations, where uncertain treatment
effects are characterized by interval-valued probabilities. An attention
mechanism with learnable query-key transformations enables flexible,
data-driven aggregation of pairwise comparisons, while simultaneously learning
soft class probabilities for censored cases. Through extensive experiments on
synthetic datasets with complex nonlinear structures, including spiral,
bell-shaped, and circular feature spaces, we demonstrate that our approach
maintains robust performance across varying censoring rates and treatment
effect strengths. The model consistently outperforms meta-learner baselines
(T-learner and S-learner) equipped with random survival forests, Cox
proportional hazards, and Beran estimators, particularly in challenging
nonlinear scenarios where conventional methods exhibit significant
degradation. The results establish the proposed attention-based framework as a
scalable and statistically principled solution for personalized treatment
benefit assessment in survival settings. The code implementing
the model is publicly available.


\end{abstract}

\section{Introduction}

Clinically grounded medicine has historically depended on the \emph{average
treatment effect (ATE)} to assess the effectiveness of therapeutic
interventions. This approach aggregates data from entire populations to
classify whether a treatment is better, comparable, or worse than a control
for an average patient, thereby supporting regulatory decision-making.
Nevertheless, this population-focused perspective faces limitations when it
comes to clinical practice, where decisions are primarily driven by
predictions tailored to individual patients.

In the context of survival analysis, patient outcomes are expressed as the
time until the occurrence of an event of interest such as death, disease
recurrence, or relapse. The primary clinical concern is not merely whether a
treatment affects some average time-to-event but rather the probability that,
for a specific patient, the treatment will extend or improve their individual
outcome. This leads us to consider the \emph{Individual Probability of
Treatment Benefit (IPTB)} in survival settings, which quantifies the
likelihood that the potential survival time under treatment exceeds that under
control, given patient-specific covariates.

Formally, let $H$ and $Y$ denote the potential survival times under treatment
and control, respectively, for a patient with covariates $X=\mathbf{x}$. The
IPTB is defined as the probability that the treatment results in a longer
survival time:
\begin{equation}
\rho(\mathbf{x})=\Pr \{H>Y\mid X=\mathbf{x}\}=\Pr \{ \Delta>0\mid X=\mathbf{x}
\},
\end{equation}
where $\Delta= H - Y$ is the individual treatment effect in terms of time to event.

Estimating $\Pr \{ \Delta>0\mid X=\mathbf{x}\}$ enables a more nuanced
understanding of treatment benefits tailored to individual patients. Unlike
the Conditional Average Treatment Effect (CATE), which estimates average
differences in survival times across groups, IPTB directly measures the
probability that a particular patient will experience a longer survival time
under treatment. This patient-specific approach aligns with clinical
decision-making, accommodates individual variability in biological and
lifestyle factors, and addresses the limitations of subgroup averages which
often do not translate into accurate individual risk assessments
\cite{kent2020predictive,rekkas2020predictive,ruiz2022non}.

From a statistical perspective, deriving IPTB involves reconstructing 
the joint distribution of potential survival times under the fundamental causal inference
challenge that only one potential outcome can be observed for each patient. 
This necessitates modeling the full conditional
distributions of survival times given covariates, which can be approached
effectively through Bayesian methods or advanced semi-parametric and
frequentist techniques \cite{Fava-25}.

In this work, we focus on developing a framework for estimating the IPTB
within survival analysis, explicitly addressing the complexities introduced by
censored data common in survival studies where the event of interest may not
have occurred for some patients by the study end. Handling censored
observations is essential to avoid biased estimates of survival probabilities
and treatment effects. Standard methods like the Kaplan-Meier estimator and
Cox proportional hazards models provide essential tools for survival data, but
estimating the probability that a treatment improves an individual survival
time requires further extension to incorporate covariate-dependent
distributions and censored information. The proposed model is called Surv-IPTB. 

We recast the estimation of $\Pr \{ \Delta>0\mid X=\mathbf{x}\}$ as a binary
classification problem, where the classes correspond to whether the difference
in survival times is positive or negative. This involves analyzing paired
observations $(h_{i}, y_{j})$, representing potential survival times under
treatment and control for different patients, and constructing non-parametric,
interval-based probability distributions that accommodate prior uncertainty,
especially for censored observations. To synthesize these distributions
conditioned on covariates, we employ feature-dependent aggregation models,
such as Nadaraya-Watson kernel regression \cite{Nadaraya-1964,Watson-1964} and
attention mechanisms \cite{Luong-etal-2015,Vaswani-etal-17}. This approach
extends recent advancements in survival analysis
\cite{Konstantinov-Utkin-etal-25}.

Our key contributions are as follows:

\begin{enumerate}
\item We propose a novel model for estimating the IPTB in survival contexts
with \textquotedblleft soft\textquotedblright \ parametric assumptions about
the distributions of survival times. The model employs attention mechanisms
and can be extended to more complex architectures, including transformers. In
addition, the model avoids ambiguity in determining the treatment effect for
censored data.

\item Through extensive experiments with both synthetic and semi-synthetic
datasets, we demonstrate the effectiveness of the proposed methods in
capturing individual treatment benefits amid the challenges posed by censored
data. Our experiments encompass diverse data structures, including, linear,
bell-shaped, circular, and spiral feature spaces, representing varying degrees
of complexity and nonlinearity. We provide a rigorous comparative analysis
against six meta-learner baselines (T-learner and S-learner combined with
random survival forests, Cox proportional hazards, and Beran estimators),
demonstrating that our approach maintains superior performance across
challenging scenarios where conventional methods exhibit significant
degradation. We investigate the sensitivity of our model to key parameters
including censoring rate, treatment power, and treatment group size, providing
insights into its robustness and practical applicability.

The implementation code is publicly available at:
{\url{https://github.com/NTAILab/SurvIPTB/tree/main}}.
\end{enumerate}

The paper is organized as follows. Section 2 reviews related work on CATE
estimation, survival analysis with treatment effect heterogeneity, and IPTB
estimation. Section 3 formalizes the problem setup, introduces notation, and
presents the attention-based estimation framework with detailed treatment of
censoring scenarios. Section 4 describes our experimental methodology,
including synthetic data generation, benchmark models, and validation schemes.
Section 5 presents comprehensive results across all datasets and scenarios,
with detailed analysis and discussion. Section 6 concludes the paper and
outlines directions for future research.

\section{Related Work}

\textbf{Conditional Average Treatment Effect Estimation.} CATE estimation has
evolved considerably over the past decade. Early approaches relied on
regularized regression models and support vector machines
\cite{Jeng-Lu-Peng-2018,Zhou-Mayer-Hamblett-etal-2017}, with subsequent
extensions to censored data \cite{zhang2017mining}. Bayesian nonparametric
methods \cite{Alaa-Schaar-2018}, multiple testing formulations
\cite{Xie-Chen-Shi-2018}, and ensemble-based evaluations
\cite{Wendling-etal-2018} further expanded the methodological toolkit.
Comprehensive reviews provide detailed overviews of available techniques
\cite{Caron-etal-22,curth2024using,yao2021survey,Zhang-Li-Liu-22}.

Recent advances have introduced more flexible and scalable approaches.
Meta-learners, including T-, S-, and X-learners, enable principled combination
of base estimators \cite{Wang-etal-2016,Kunzel-etal-2018}. Deep learning
methods capture complex functional relationships in treatment effect
heterogeneity
\cite{Bica-etal-20,Curth-Schaar-21a,Nie-etal-21,Qin-Wang-Zhou-21,shi2024estimating}%
, while transformer-based architectures offer an alternative for
representation learning in this context
\cite{Guo-Zheng-etal-21,Melnychuk-etal-22,zhang2023exploring}.

\textbf{Survival Analysis with Treatment Effect Heterogeneity.} Extending
treatment effect estimation to survival outcomes requires careful treatment of
censoring. Foundational approaches include nonparametric least squares for
survival times \cite{park2013estimation} and causal survival forests tailored
to right-censored outcomes \cite{cui2023estimating}. Meta-learning frameworks
adapted for survival analysis employ survival-specific base learners such as
random survival forests and Bayesian AFT models \cite{bo2024meta}, while
pseudo individualized treatment effects provide an alternative basis for HTE
quantification \cite{bo2025evaluating}.

Hybrid architectures combine established techniques: Counterfactual Survival
Analysis has been integrated with Dragonnet-style representations
\cite{zhao2022estimating}, and Bayesian additive regression trees have been
paired with random-intercept AFT models \cite{hu2024new}. Orthogonal learning
principles extend to the survival setting to address censoring-induced bias
\cite{frauen2025orthogonal}. Nonparametric AFT models provide flexible
alternatives for individualized HTE estimation
\cite{henderson2020individualized}. Additional contributions to survival HTE
include
\cite{Ozenne-etal-2020,schrod2022bites,Trinquart-etal-16,zhang2017mining,Zhu-Gallego-20}%
.

\textbf{Individual Probability of Treatment Benefit in Survival Settings.}
Estimating whether a treatment improves individual survival time, which is a
key component of IPTB in survival analysis, requires moving beyond average
treatment effects to characterize the full distribution of treatment benefit
conditional on patient covariates. This task involves constructing
non-parametric, interval-based probability distributions that accommodate
censoring-related uncertainty, since observed survival times are often
incomplete due to study follow-up limitations.

Existing IPTB approaches employ Bayesian modeling of joint potential outcome
distributions \cite{zhang2023exploring} and probabilistic classification
schemes that directly model the benefit function $\Delta=H-Y$
\cite{Melnychuk-etal-22,melnychuk2024quantifying}. However, adapting these
methods to survival outcomes demands careful treatment of censored
observations to avoid biased estimates of whether an individual would benefit.

IPTB estimation in survival contexts inherits the fundamental challenges from
CATE while introducing additional complexities. Beyond the unobserved
counterfactual problem, characterizing the probability that treatment improves
an individual's survival time requires methods that: (i) properly incorporate
censoring information rather than ignoring incomplete observations, (ii)
construct covariate-dependent distributions over paired survival times under
different treatment regimes, and (iii) employ sufficiently flexible
aggregation models to capture heterogeneous treatment benefit patterns. These
requirements demand stronger statistical assumptions and more sophisticated
machinery than estimating conditional mean effects alone.

\section{Problem Formulation and Notation}

We consider an observational study with two treatment groups: control and
treatment. The control group is represented by dataset $\mathcal{D}%
_{0}=\left \{  (\mathbf{x}_{i},y_{i})\right \}  _{i=1}^{c}$, comprising $c$
independent observations. For each subject $i$, $\mathbf{x}_{i}\in
\mathbb{R}^{d}$ denotes a $d$-dimensional covariate vector capturing patient
characteristics, and $y_{i}\in \mathbb{R}$ represents the observed outcome
under control (e.g., survival time or clinical measurement). Correspondingly,
the treatment group $\mathcal{D}_{1}=\left \{  (\mathbf{z}_{j},h_{j})\right \}
_{j=1}^{t}$ contains $t$ subjects receiving active treatment. Here,
$\mathbf{z}_{j}\in \mathbb{R}^{d}$ is the covariate vector and $h_{j}%
\in \mathbb{R}$ the observed outcome under intervention.

We adopt the potential outcomes framework \cite{Rubin-2005} to formalize
causal inference. For each subject, let $Y$ denote the potential outcome under
control ($A=0$) and $H$ denote the potential outcome under treatment ($A=1$).
These counterfactual quantities represent what would occur under each
treatment regime, though in practice only one is observed.

Our inferential goal departs from classical CATE estimation, which quantifies
the conditional average treatment effect
\begin{equation}
\tau(\mathbf{x})=\mathbb{E}[H-Y\mid \mathbf{X}=\mathbf{x}].
\end{equation}

Instead, we focus on IPTB, defined as the probability that an individual with
covariate profile $\mathbf{X}=\mathbf{x}$ experiences a strictly positive
treatment effect:
\begin{equation}
\rho(\mathbf{x})=\Pr \left \{  H>Y\mid \mathbf{X}=\mathbf{x}\right \}  .
\end{equation}

This quantity provides a direct probabilistic assessment of whether treatment
improves outcomes at the individual level. Unlike the conditional average
effect, $\rho(\mathbf{x})$ captures the full heterogeneity in treatment
response across the outcome distribution, revealing whether treatment is
beneficial (probability exceeding 0.5), harmful (probability below 0.5), or
neutral (probability near 0.5) for a given patient profile. This
characterization is particularly valuable in clinical settings where
decision-makers seek to identify patients likely to benefit from intervention,
rather than only quantifying average improvements.

\subsection{Formal Problem Statement for Survival Data}

In the specific context of survival analysis, the outcomes $Y$ and $H$
represent latent event times (e.g., time to death or disease progression).
However, due to the finite duration of studies or loss to follow-up, these
event times are subject to right-censoring. We formally extend the data
structures $\mathcal{D}_{0}$ and $\mathcal{D}_{1}$ to account for censoring indicators.

Let $Y^{\ast}$ and $H^{\ast}$ denote the true latent survival times under
control and treatment, respectively. Let $C^{0}$ and $C^{1}$ denote the
corresponding censoring times. The observed outcomes and censoring statuses
for the control group are defined as:
\begin{equation}
y_{i}=\min(Y_{i}^{\ast},C_{i}^{0}),\quad \xi_{i}=\mathbb{I}(Y_{i}^{\ast}\leq
C_{i}^{0}),
\end{equation}
where $\xi_{i}\in \{0,1\}$ is the event indicator ($\xi_{i}=1$ if the event is
observed, $\xi_{i}=0$ if censored). Similarly, for the treatment group:
\begin{equation}
h_{j}=\min(H_{j}^{\ast},C_{j}^{1}),\quad \delta_{j}=\mathbb{I}(H_{j}^{\ast}\leq
C_{j}^{1}),
\end{equation}
where $\delta_{j}\in \{0,1\}$ is the treatment event indicator. Consequently,
the observed datasets are triplets:
\begin{equation}
\mathcal{D}_{0}=\left \{  (\mathbf{x}_{i},y_{i},\xi_{i})\right \}  _{i=1}%
^{c},\  \mathcal{D}_{1}=\left \{  (\mathbf{z}_{j},h_{j},\delta_{j})\right \}
_{j=1}^{t}.
\end{equation}

A fundamental concept in survival analysis is the survival function (SF)
$S(t\mid \mathbf{x})$, defined as the probability that an individual survives
beyond time $t$. We denote SFs trained on $\mathcal{D}_{0}$ and $\mathcal{D}%
_{1}$ as $S_{0}(t\mid \mathbf{x})$ and $S_{1}(t\mid \mathbf{z})$, respectively.

Given the censored observational datasets $\mathcal{D}_{0}$ and $\mathcal{D}%
_{1}$, and assuming standard causal identifiability conditions (consistency,
conditional ignorability, and positivity), the objective is to construct an
estimator $\widehat{\Psi}(\mathbf{x})$ for the Individual Probability of
Treatment Benefit (IPTB):
\begin{equation}
\Psi(\mathbf{x})=\Pr \left(  H^{\ast}>Y^{\ast}\mid \mathbf{X}=\mathbf{x}\right)
,
\end{equation}
without direct access to the uncensored latent variables $Y^{\ast}$ and
$H^{\ast}$. The estimator must account for the informative nature of the
censoring mechanisms and the potential selection bias inherent in the
observational assignment $A$.

Under assumptions of conditional exchangeability and positivity \cite{Rubin-2005}, the observed
data identify conditional marginal distributions $F_{H^{\ast}}(t\mid \mathbf{x}%
)=\Pr \{H^{\ast} \leq t\mid X=\mathbf{x}\}$ and $F_{Y^{\ast}}(t\mid \mathbf{x})=\Pr \{Y^{\ast} \leq t\mid X=\mathbf{x}\}$. However, the joint conditional distribution of the two
potential outcomes is not identified. We can represent it using a copula
\cite{nelsen2006introduction}:
\[
F_{H^{\ast},Y^{\ast}}(h,y\mid \mathbf{x})=C_{\theta,\mathbf{x}}\left(  F_{H^{\ast}}(h\mid
\mathbf{x}),F_{Y^{\ast}}(y\mid \mathbf{x})\right)  .
\]

Here the copula $C_{\theta,\mathbf{x}}$ defines how two potential outcomes are
related in the same patient; $\theta$ is the dependence parameter. The current
pairwise construction corresponds to the independence copula $C(u,v)=u\cdot
v$. Because the dependence parameter $\theta$ is not identified from
parallel-arm observational data, alternative values can be examined through
sensitivity analysis.

\section{Treatment Benefit Probability as a Binary Classification Framework}

\subsection{Pairwise Patient Comparison and Censoring Structure}

The core methodology involves constructing pairwise comparisons between
patients from distinct groups. Specifically, we form pairs consisting of one
treatment group patient $(\mathbf{z}_{i},h_{i},\delta_{i})$ and one control
group patient $(\mathbf{x}_{j},y_{j},\xi_{j})$. The treatment effect for each
pair is quantified as $\Delta_{ij}=h_{i}-y_{j}$.

The censoring status of $\Delta_{ij}$ is determined by the censoring
indicators $\delta_{i}$ and $\xi_{j}$, yielding four distinct scenarios
corresponding to the positive treatment effect:

\begin{description}
\item[Case 1:] Both treatment and control observations are uncensored
($\delta_{i}=1$ and $\xi_{j}=1$), and $h_{i}-y_{j}>0$. The treatment effect
$\Delta_{ij}$ is fully observed and positive.

\item[Case 2:] Treatment observation is censored while control is uncensored
($\delta_{i}=0$ and $\xi_{j}=1$), and $h_{i}-y_{j}>0$. The treatment effect
undergoes left censoring.

\item[Case 3:] Treatment observation is censored while control is uncensored
($\delta_{i}=0$ and $\xi_{j}=1$), and $h_{i}-y_{j}\leq0$. The treatment effect
undergoes right censoring.

\item[Case 4:] Treatment observation is uncensored while control is censored
($\delta_{i}=1$ and $\xi_{j}=0$), and $h_{i}-y_{j}>0$. The treatment effect
undergoes left censoring.
\end{description}

\begin{figure}
[h!]
\begin{center}
\includegraphics[
height=2.7363in,
width=5.1015in
]%
{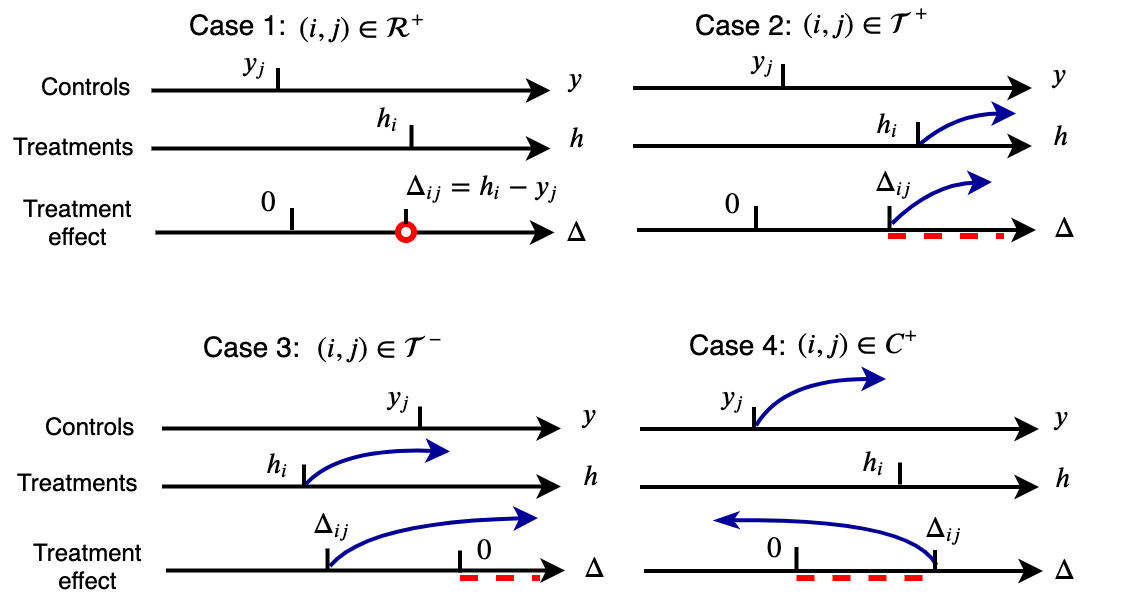}%
\caption{Four cases of subsets where $\Delta>0$}%
\label{f:explan_iptb_pos}%
\end{center}
\end{figure}

In the case when both observations are censored ($\delta_{i}=0$ and $\xi
_{j}=0$), the treatment effect is completely indeterminate and excluded from
analysis. These cases are depicted in Fig. \ref{f:explan_iptb_pos}.%

\begin{figure}
[h!]
\begin{center}
\includegraphics[
height=2.8522in,
width=5.1586in
]%
{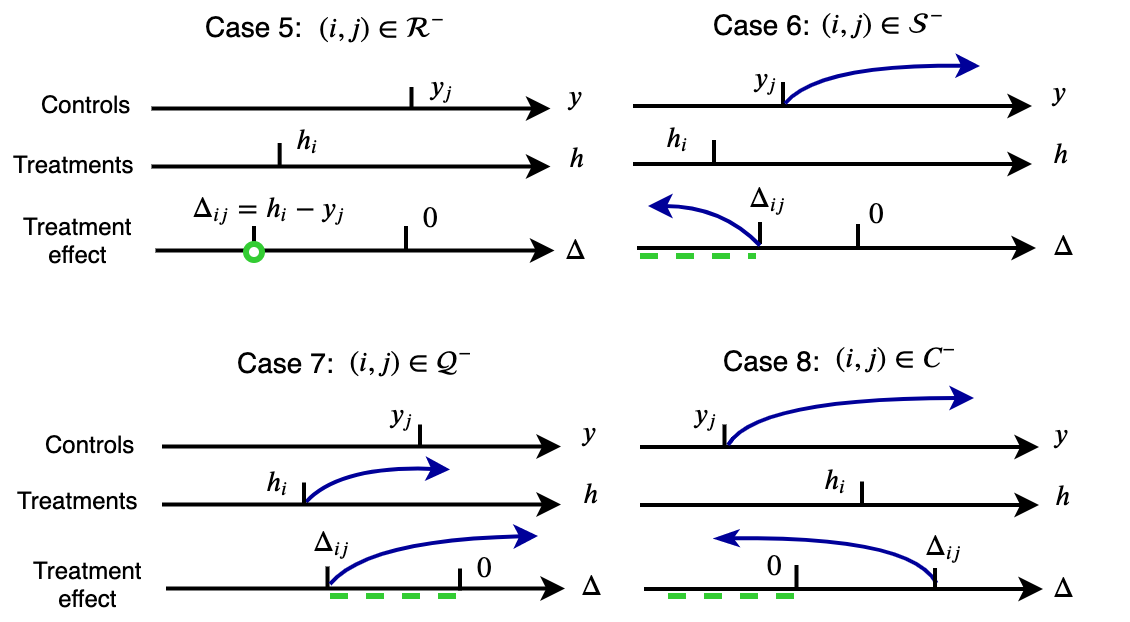}%
\caption{Four cases of subsets where $\Delta \leq0$}%
\label{f:explan_iptb_negat}%
\end{center}
\end{figure}

In the same way, we consider cases of the negative treatment effect (see Fig.
\ref{f:explan_iptb_negat}):

\begin{description}
\item[Case 5:] Both treatment and control observations are uncensored
($\delta_{i}=1$ and $\xi_{j}=1$), and $h_{i}-y_{j}\leq0$. The treatment effect
$\Delta_{ij}$ is fully observed and negative.

\item[Case 6:] Treatment observation is uncensored while control is censored
($\delta_{i}=1$ and $\xi_{j}=0$), and $h_{i}-y_{j}\leq0$. The treatment effect
undergoes left censoring.

\item[Case 7:] Treatment observation is censored while control is uncensored
($\delta_{i}=0$ and $\xi_{j}=1$), and $h_{i}-y_{j}\leq0$. The treatment effect
undergoes right censoring.

\item[Case 8:] Treatment observation is uncensored while control is censored
($\delta_{i}=1$ and $\xi_{j}=0$), and $h_{i}-y_{j}>0$. The treatment effect
undergoes left censoring.
\end{description}

%

Cases 3 and 4 coincide with Cases 7 and 8, respectively, but they consider
different regions of $\Delta_{ij}$.

\subsection{Treatment Effect Partitioning}

We collect all ordered treatment effect values $\Delta_{ij}$ for $i=1,\dots,t$
and $j=1,\dots,c$, into two subsets $\mathcal{G}^{+}$ and $\mathcal{G}^{-}$
with positive and negative values $\Delta_{ij}$, respectively, i.e., we can
write
\begin{equation}
\mathcal{G}^{+}=\{ \Delta_{ij}:\Delta_{ij}>0\},\  \mathcal{G}^{-}=\{
\Delta_{ij}:\Delta_{ij}\leq0\}.
\end{equation}

The values of $\Delta_{ij}$ define the class values in the binary
classification task which are between $0$ and $1$. For some cases of the
censored and uncensored observations described above, we cannot assign class
values to $0$ or $1$. For example, the class values in Cases 3 and 4 are
uncertain. Using the machine learning terminology of multi-class
classification tasks, we will say about class probabilities rather than class values.

Let us consider subsets of $\Delta_{ij}$ whose values can belong to the
interval $(0,+\infty)$, and introduce the following sets of indices
corresponding to these subsets:

\begin{itemize}
\item The index set $\mathcal{R}^{+}$ corresponding to all pairs of uncensored
observations ($\delta_{i}=1$ and $\xi_{j}=1$) with positive values
$\Delta_{ij}$ (from $\mathcal{G}^{+}$):
\begin{equation}
\mathcal{R}^{+}=\{(i,j):\Delta_{ij}>0,\delta_{i}=1,\xi_{j}=1\}.
\end{equation}

This case is simple because $\Delta_{ij}>0$ and the class probability in the
binary classification task is $1$.

\item The index set $\mathcal{T}^{+}$ corresponding to all pairs of
observations with $\delta_{i}=0$, $\xi_{j}=1$ under condition $\Delta_{ij}>0$:%
\begin{equation}
\mathcal{T}^{+}=\{(i,j):\Delta_{ij}>0,\delta_{i}=0,\xi_{j}=1\}.
\end{equation}
In this case, due to right censoring, we do not know when the event precisely
occurs, but we exactly know that it will be in the positive part of values
$\Delta_{ij}$. This implies that the class probability in the binary
classification task is also $1$.

\item The index set $\mathcal{T}^{-}$ corresponding to all pairs of
observations with $\delta_{i}=0$ and $\xi_{j}=1$ under condition $\Delta
_{ij}\leq0$:
\begin{equation}
\mathcal{T}^{-}=\{(i,j):\Delta_{ij}\leq0,\delta_{i}=0,\xi_{j}=1\}.
\end{equation}
This set contains indices of all right-censored observations $\Delta_{ij}$.
Due to right censoring, the set of possible $\Delta_{ij}$ contains positive
and negative values. This case is more complex. The problem is that we do not
know precise value $\Delta_{ij}$ of the event, and we do not know wether it
positive or negative. This leads to the imprecise classprobability that the
event will be in the positive part, i.e., it is in the interval $[0,1]$. This
class probability may be arbitrary in the interval from 0 to 1. The interval
$[0,1]$ reflects epistemic uncertainty regarding probabilities of classes.

\item The index set $\mathcal{C}^{+}$ corresponding to all pairs of
observations with $\delta_{i}=1$ and $\xi_{j}=0$ under condition that
$\Delta_{ij}>0$:
\begin{equation}
\mathcal{C}^{+}=\{(i,j):\Delta_{ij}>0,\delta_{i}=1,\xi_{j}=0\}.
\end{equation}
In this case, we have left-cenored observations $\Delta_{ij}$. We again do not
know in which part of $\Delta$ will be event, but we know that there is
possibility that event will be in positive part but with unknown class
probability in the interval $[0,1]$.
\end{itemize}

The relationship between the outcomes $h_{i}$ and $y_{j}$ and their censoring
indicators defines sets of possible values of $\Delta$ for which the condition
$\Delta>0$ is satisfied. These regions are indicated by the red dotted lines
and arise due to censored observations, for which the exact event time is
unknown. As shown in Fig. \ref{f:explan_iptb_pos} (Case 2), when $\Delta
_{ij}>0$, all points within the interval $[\Delta_{ij},\infty)$ are greater
than zero. In Case 3, $\Delta_{ij}$ can be negative, however, due to the
censoring of a treated patient, a possible event time may lie within the
interval $(0,\infty)$. It could also lie within the interval $[\Delta_{ij},0]$
with some unknown probability. This implies that the probability of the event
occurring within $(0,\infty)$ is also unknown. The same reasoning applies to
Case 4.

\subsection{Challenge in Cases 3, 4, 7, and 8}

By treating the IPTB problem as a binary classification task, we can precisely
determine the class label for any pair $(\mathbf{z}_{i},\mathbf{x}_{j})$ with
$(i,j)\in \mathcal{R}^{+}\cup \mathcal{T}^{+}$, corresponding to Cases 1 and 2.
In contrast, the primary challenge in Cases 3 and 4 is that $\Delta_{ij}$ can
be either positive or negative, and we do not know the true class labels or,
at minimum, their probabilities (soft label definition) required to define the
loss function. A straightforward approach would be to consider only pairs
$(\mathbf{z}_{i},\mathbf{x}_{j})$ with $(i,j)\in \mathcal{R}^{+}\cup
\mathcal{T}^{+}$. However, censored observations are often numerous, and
discarding pairs $(\mathbf{z}_{i},\mathbf{x}_{j})$ with $(i,j)\in
\mathcal{T}^{-} \cup \mathcal{C}^{+}$ results in loss of valuable information.
Therefore, we propose the following approach to leverage these pairs: we
estimate soft class probabilities $\phi_{i,j}$ for $(i,j)\in \mathcal{T}^{-}$
(Case 3) and $\rho_{i,j}$ for $(i,j)\in \mathcal{C}^{+}$ (Case 4). The
corresponding negative class probabilities are defined as $1-\phi_{i,j}$ and
$1-\rho_{i,j}$, respectively.

To derive the probability $\phi_{i,j}$, we consider the random treatment event
time $H^{\ast}$. For each pair $(\mathbf{z}_{i},\mathbf{x}_{j})$, the positive
region of $\Delta_{ij}$ in Case 3 corresponds to the event $H^{\ast}\geq
y_{j}$ conditional on $H^{\ast}>h_{i}$. Thus, $\phi_{i,j}$ is defined as
\begin{align}
\phi_{i,j}  &  =\Pr \left \{  H^{\ast}\geq y_{j}\mid H^{\ast}>h_{i}\right \}
=\frac{\Pr \left \{  (H^{\ast}\geq y_{j})\cap(H^{\ast}>h_{i})\right \}  }%
{\Pr \left \{  H^{\ast}>h_{i}\right \}  }\nonumber \\
&  =\frac{\Pr \left \{  H^{\ast}\geq y_{j}\right \}  }{\Pr \left \{  H^{\ast}%
>h_{i}\right \}  }~\text{for }h_{i}\leq y_{j}.
\end{align}

In terms of SFs, the probability $\phi_{i,j}$ is expressed through the SF
$S_{1}(t)$ trained on the treatment group $\mathcal{D}_{1}=\left \{
(\mathbf{z}_{j},h_{j},\delta_{j})\right \}  _{j=1}^{t}$ as follows:
\begin{equation}
\phi_{i,j}=\frac{S_{1}(y_{j})}{S_{1}(h_{i})}.
\end{equation}

The SF $S_{1}(t)$ can be estimated using the Kaplan-Meier estimator or other
methods such as the Beran estimator or Cox proportional hazards model.

Similarly, the probability $\rho_{i,j}$ for Case 4 can be derived. We consider
the random control event time $Y^{\ast}$. For each pair $(\mathbf{z}%
_{i},\mathbf{x}_{j})$, the positive region of $\Delta_{ij}$ in Case 4
corresponds to the event $y_{j}\leq Y^{\ast}\leq h_{i}$ conditional on
$Y^{\ast}\geq y_{j}$. Thus, $\rho_{i,j}$ is defined as
\begin{align}
\rho_{i,j}  &  =\Pr \left \{  y_{j}\leq Y^{\ast}\leq h_{i}\mid Y^{\ast}\geq
y_{j}\right \}  =\frac{\Pr \left \{  (y_{j}\leq Y^{\ast}\leq h_{i})\cap(Y^{\ast
}\geq y_{j})\right \}  }{\Pr \left \{  Y^{\ast}\geq y_{j}\right \}  }\nonumber \\
&  =\frac{\Pr \left \{  y_{j}\leq Y^{\ast}\leq h_{i}\right \}  }{\Pr \left \{
Y^{\ast}\geq y_{j}\right \}  }~\text{for }y_{j}\leq h_{i}.
\end{align}

In terms of survival functions, the probability $\rho_{i,j}$ is expressed
through the SF $S_{0}(t)$ estimated from the control group $\mathcal{D}_{0}$
as follows:
\begin{equation}
\rho_{i,j}=\frac{S_{0}(y_{j})-S_{0}(h_{i})}{S_{0}(y_{j})}.
\end{equation}

We now consider Cases 5--8, which correspond to negative treatment effects and
are depicted in Fig. \ref{f:explan_iptb_negat}. The analysis of these cases
parallels that of Cases 1--4. Cases 5 and 6 are characterized by precisely
known class labels, whereas Cases 7 and 8 coincide with Cases 3 and 4 but with
negative treatment effect regions. These cases have probabilities
$1-\phi_{i,j}$ and $1-\rho_{i,j}$ for Cases 7 and 8, respectively. These
probabilities will be incorporated into the loss function.

It is important to note that we have considered unconditional SFs that do not
depend on $\mathbf{z}_{i}$ and $\mathbf{x}_{j}$. The reason is that computing
conditional SFs for these cases is a rather complex problem, which reduces to
multi-class classification and the additional use of survival models
that account for conditions on $\mathbf{z}_{i}$ and $\mathbf{x}_{j}$. To
simplify the computations, we have proposed an approximate simple approach,
the effectiveness of which is confirmed by numerical experiments.

So, our aim is to find the probability that the treatment effect will be
positive for a new pair of patients with feature vectors $\mathbf{z}$ and
$\mathbf{x}$, that is,
\begin{equation}
\Pr \{ \Delta>0\mid \mathbf{Z}=\mathbf{z},\mathbf{X}=\mathbf{x}\}.
\end{equation}

\subsection{Attention-based estimation}

Let $p^{+}(\mathbf{z},\mathbf{x})$ denote the probability that the difference
between observations with treatment instance $\mathbf{z}$ and control instance
$\mathbf{x}$ yields $\Delta$ belonging to one of the sets $\mathcal{R}^{+}$,
$\mathcal{T}^{+}$, or $\mathcal{C}^{+}$. Then we can write
\begin{align}
p^{+}(\mathbf{z},\mathbf{x})  &  =\sum_{(i,j)\in \mathcal{R}^{+}}%
a(\mathbf{z},\mathbf{x},\mathbf{z}_{i},\mathbf{x}_{j})\cdot1\nonumber \\
&  +\sum_{(l,k)\in \mathcal{T}^{+}}a(\mathbf{z},\mathbf{x},\mathbf{z}%
_{l},\mathbf{x}_{k})\cdot1\nonumber \\
&  +\sum_{(v,w)\in \mathcal{T}^{-}}a(\mathbf{z},\mathbf{x},\mathbf{z}%
_{v},\mathbf{x}_{w})\cdot \lbrack0,1]\nonumber \\
&  +\sum_{(r,s)\in \mathcal{C}^{+}}a(\mathbf{z},\mathbf{x},\mathbf{z}%
_{r},\mathbf{x}_{s})\cdot \lbrack0,1].
\end{align}

The attention weights are defined via kernel functions:
\begin{equation}
a(\mathbf{z},\mathbf{x},\mathbf{z}_{i},\mathbf{x}_{j})=\frac{K(\mathbf{z}%
,\mathbf{z}_{i})\cdot K(\mathbf{x},\mathbf{x}_{j})}{\sum_{(r,s)\in \mathcal{G}%
}K(\mathbf{z},\mathbf{z}_{r})\cdot K(\mathbf{x},\mathbf{x}_{s})},
\end{equation}
where $\mathcal{G}$ denotes the set of all possible indices $(r,s)$ of feature
vectors $(\mathbf{z}_{r},\mathbf{x}_{s})$.

Intervals $[0,1]$ in the expression for $p^{+}(\mathbf{z},\mathbf{x})$
indicate that we do not know probabilities that difference of outcomes for the
pair $(\mathbf{z},\mathbf{x})$ is positive. Denote the probabilities $\Pr \{
\Delta>0\mid \mathbf{Z}=\mathbf{z}_{i},\mathbf{X}=\mathbf{x}_{j}\}$ for
$(i,j)\in \mathcal{T}^{-}$ and $(i,j)\in \mathcal{C}^{+}$ as $\pi^{(i,j)}$. Then
the probability $p^{+}(\mathbf{z},\mathbf{x})$ can be rewritten as follows:
\begin{align}
p^{+}(\mathbf{z},\mathbf{x})  &  =\sum_{(i,j)\in \mathcal{R}^{+}\cup
\mathcal{T}^{+}}a(\mathbf{z},\mathbf{x},\mathbf{z}_{i},\mathbf{x}%
_{j})\nonumber \\
&  +\sum_{(r,s)\in \mathcal{T}^{-}\cup \mathcal{C}^{+}}a(\mathbf{z}%
,\mathbf{x},\mathbf{z}_{r},\mathbf{x}_{s})\cdot \pi^{(r,s)},
\end{align}
where $\sum_{(r,s)\in \mathcal{G}}a(\mathbf{z},\mathbf{x},\mathbf{z}%
_{r},\mathbf{x}_{s})=1$.

Here $\pi^{(r,s)}\in \lbrack0,1]$ represents the imprecise probability that the
censored event $\Delta_{rs}$ occurs in the interval $(0,\infty)$ for
$(r,s)\in \mathcal{T}^{-}$ or in the interval $(0,\Delta_{rs})$ for
$(r,s)\in \mathcal{C}^{+}$.

The probability $p^{-}(\mathbf{z},\mathbf{x})=\Pr \{ \Delta<0\mid
\mathbf{Z}=\mathbf{z},\mathbf{X}=\mathbf{x}\}$ equals $1-p^{+}(\mathbf{z}%
,\mathbf{x})$.

Following the attention mechanism framework
\cite{Luong-etal-2015,Vaswani-etal-17}, we treat $(\mathbf{z},\mathbf{x})$ as
the \emph{query}, $(\mathbf{z}_{i},\mathbf{x}_{j})$ as \emph{keys}, and
$\pi^{(i,j)}$ as \emph{values}. We define:
\begin{align}
\mathbf{q}  &  =\mathbf{W}_{Q}[\mathbf{z},\mathbf{x}]^{\top}\in \mathbb{R}%
^{d},\nonumber \\
\mathbf{k}_{ij}  &  =\mathbf{W}_{K}[\mathbf{z}_{i},\mathbf{x}_{j}]^{\top}%
\in \mathbb{R}^{d},\nonumber \\
v_{ij}  &  =1\text{ or }\pi^{(i,j)}\in \lbrack0,1],
\end{align}
where $\mathbf{W}_{Q}\in \mathbb{R}^{d\times2d}$ and $\mathbf{W}_{K}%
\in \mathbb{R}^{d\times2d}$ are learnable weight matrices. The attention
weights are then:
\begin{equation}
a(\mathbf{z},\mathbf{x},\mathbf{z}_{i},\mathbf{x}_{j})=\frac{\exp \left(
\frac{1}{\sqrt{2d}}\mathbf{q}^{\top}\mathbf{k}_{ij}\right)  }{\sum
_{(s,r)\in \mathcal{G}}\exp \left(  \frac{1}{\sqrt{2d}}\mathbf{q}^{\top
}\mathbf{k}_{sr}\right)  }.
\end{equation}

Let $\mathbf{K}\in \mathbb{R}^{|\mathcal{G}|\times d}$ contain all keys and
$\mathbf{V}\in \mathbb{R}^{|\mathcal{G}|\times1}$ contain all values. The
distribution can be expressed in matrix form:
\begin{equation}
p^{+}(\mathbf{z},\mathbf{x})=\text{softmax}\left(  \frac{1}{\sqrt{2d}%
}\mathbf{q}\mathbf{K}^{\top}\right)  \mathbf{V}\in \mathbb{R}.
\end{equation}

Here a part of the vector $\mathbf{V}=(1,...,1,\pi^{(i,j)},...,\pi^{(r,s)})$
is equal to $1$ and is not trained.

The model learns matrices $\mathbf{W}_{Q}$ and $\mathbf{W}_{K}$, and
probabilities $\pi^{(r,s)}$ by minimizing the log-likelihood loss:
\begin{align}
\mathcal{L}(\mathbf{p}  &  \mid \mathbf{W}_{Q},\mathbf{W}_{K})=L(\mathcal{R}%
^{+}\cup \mathcal{T}^{+})+L(\mathcal{R}^{-}\cup \mathcal{S}^{-})\nonumber \\
&  +L(\mathcal{T}^{-})+L(\mathcal{C}^{+})+L(\mathcal{Q}^{-})+L(\mathcal{C}%
^{-})+\gamma L(\mathbf{\pi})+\eta L(\mathbf{W}),
\end{align}
where $L(\mathcal{R}^{+}\cup \mathcal{T}^{+})$ is the log-likelihood loss for
pairs $(\mathbf{z}_{i},\mathbf{x}_{j})$ with $\Delta_{ij}>0$, defined as
\begin{equation}
L(\mathcal{R}^{+}\cup \mathcal{T}^{+})=-\sum_{(i,j)\in \mathcal{R}^{+}%
\cup \mathcal{T}^{+}}\log \left(  p^{+}(\mathbf{z}_{i},\mathbf{x}_{j})\right)  ,
\end{equation}
$L(\mathcal{R}^{-}\cup \mathcal{S}^{-})$ is the log-likelihood loss for pairs
$(\mathbf{z}_{i},\mathbf{x}_{j})$ with $\Delta_{ij}<0$, defined as
\begin{equation}
L(\mathcal{R}^{-}\cup \mathcal{S}^{-})=-\sum_{(i,j)\in \mathcal{R}^{-}%
\cup \mathcal{S}^{-}}\log \left(  p^{-}(\mathbf{z}_{i},\mathbf{x}_{j})\right)  ,
\end{equation}
$L(\mathcal{T}^{-})$, $L(\mathcal{C}^{+})$, $L(\mathcal{Q}^{-})$,
$L(\mathcal{C}^{-})$ are the log-likelihood loss functions for pairs
$(\mathbf{z}_{i},\mathbf{x}_{j})$ with soft class label probabilities, defined
as
\begin{equation}
L(\mathcal{T}^{-})=-\sum_{(r,s)\in \mathcal{T}^{-}}\phi_{rs}\log \left(
p^{+}(\mathbf{z}_{r},\mathbf{x}_{s})\right)  ,
\end{equation}%
\begin{equation}
L(\mathcal{C}^{+})=-\sum_{(r,s)\in \mathcal{C}^{+}}\rho_{rs}\log \left(
p^{+}(\mathbf{z}_{r},\mathbf{x}_{s})\right)  ,
\end{equation}%
\begin{equation}
L(\mathcal{Q}^{-})=-\sum_{(r,s)\in \mathcal{Q}^{-}}(1-\phi_{rs})\log \left(
p^{-}(\mathbf{z}_{r},\mathbf{x}_{s})\right)  ,
\end{equation}%
\begin{equation}
L(\mathcal{C}^{-})=-\sum_{(r,s)\in \mathcal{C}^{-}}(1-\rho_{rs})\log \left(
p^{-}(\mathbf{z}_{r},\mathbf{x}_{s})\right)  ,
\end{equation}
and $L(\mathbf{W})$ and $L(\mathbf{\pi})$ are regularization loss functions
defined as
\begin{equation}
L(\mathbf{W})=\Vert \mathbf{W}_{Q}\Vert^{2}+\Vert \mathbf{W}_{K}\Vert^{2},
\end{equation}%
\begin{equation}
L(\mathbf{\pi})=\sum_{(r,s)\in \mathcal{T}^{-}\cup \mathcal{C}^{+}}\pi
^{(r,s)}\log \left(  \pi^{(r,s)}\right)  ,
\end{equation}
where $\gamma$ and $\eta$ are regularization coefficients (hyperparameters).

\section{Numerical experiments}

Our evaluation is structured around two distinct validation schemes, each
designed to assess different aspects of model performance.

\paragraph{Validation scheme 1 (Val~1):}

This scheme evaluates the model's ability to estimate predicted probabilities
\begin{equation}
\Pr \{ \Delta> 0 \mid \mathbf{Z} = \mathbf{z}_{i}, \mathbf{X} = \mathbf{x}_{j}
\},
\end{equation}
for arbitrary feature pairs $(\mathbf{z}_{i}, \mathbf{x}_{j})$ drawn from test
sets. Critically, this scheme does not constrain the evaluation to matched
treatment-control pairs; rather, it assesses the model's capacity to
generalize across the entire joint feature space of treatments and controls,
thereby measuring the estimator's consistency in computing $p^{+}(\mathbf{z},
\mathbf{x})$ for any combination of treatment and control covariates.

\paragraph{Validation scheme 2 (Val~2):}

This scheme focuses specifically on matched feature vectors, where treatment
and control instances share identical covariates. For pairs $(\mathbf{x},
\mathbf{x})$, outcomes are generated according to predefined data-generating
mechanisms, enabling direct computation of
\begin{equation}
p^{+}(\mathbf{x}, \mathbf{x}) = \Pr \{ \Delta> 0 \mid \mathbf{Z} = \mathbf{x},
\mathbf{X} = \mathbf{x} \}.
\end{equation}
This scheme provides a controlled setting to evaluate performance when the
treatment and control populations are perfectly aligned in covariate space.

Model discrimination is quantified using the area under the receiver operating
characteristic curve (ROC-AUC). To ensure statistical robustness, we employ
5-fold stratified cross-validation, repeated ten times with different random
seeds, yielding stable performance estimates with associated variability
measures. Results are presented as paired ROC curves. The left panel displays
learning trajectories across Train/Val~1/Val~2 for our proposed model,
illustrating performance progression and generalization gaps. The right panel
provides comparative analysis against meta-learner baselines, including
T-learner and S-learner, each implemented with 200-tree RSF regressors, the
Cox model, the Beran estimator as base learners.

\subsection{Benchmark Models and Their Configuration}

Our comparative analysis includes Surv-IPTB and six distinct models, which are
based on using two meta-learners: the T-learner and S-learner, each paired
with three different SF estimation techniques: Random Survival Forests (RSF)
\cite{Ishwaran-Kogalur-2007}, the Cox proportional hazards model
\cite{Cox-1972}, and the Beran estimator with Gaussian kernels \cite{Beran-81}%
. The specifications of these models are detailed below in the context of
survival analysis.

\begin{enumerate}
\item \textbf{T-learner} \cite{Kunzel-etal-2018}: This approach separately
estimates the control-group SF, $S_{0}(t\mid \mathbf{x})$, and the
treatment-group SF, $S_{1}(t\mid \mathbf{x})$, for any feature vector
$\mathbf{x}$. Under some assumptions \cite{Rubin-2005}, the CATE\ can be
rewritten as:
\begin{equation}
\tau(\mathbf{x})=\mathbb{E}\left[  H\mid \mathbf{X}=\mathbf{x}\right]
-\mathbb{E}\left[  Y\mid \mathbf{X}=\mathbf{x}\right]  .
\end{equation}
Since $H$ and $Y$ are random times to events, then the CATE in the framework
of survival analysis can be computed as follows
\cite{Chapfuwa-etal-20,Trinquart-etal-16}:
\begin{equation}
\tau(\mathbf{x})=\widehat{T}_{1}(\mathbf{x})-\widehat{T}_{0}(\mathbf{x}),
\end{equation}
where $\widehat{T}_{1}(\mathbf{x})$ and $\widehat{T}_{0}(\mathbf{x})$ are
expected times to events derived from the SFs $S_{i}(t\mid \mathbf{x})$ as:
\begin{equation}
\widehat{T}_{i}(\mathbf{x})=\int_{0}^{\infty}S_{i}(t\mid \mathbf{x}%
)\mathrm{d}t,\ i=0,1.
\end{equation}

Due to the finite number of observations, the predicted SFs are step
functions, i.e., we can write
\begin{equation}
\widehat{T}_{i}(\mathbf{x})=\sum_{j=1}^{t}(t_{j}^{(i)}-t_{j-1}^{(i)})\tilde
{S}_{i}^{(j)}(\mathbf{x}),\ i=0,1,
\end{equation}
where $\tilde{S}_{i}^{(j)}(\mathbf{x})$ is the value of the SF $S_{i}%
(t\mid \mathbf{x})$ in the time interval $[t_{j-1}^{(i)},t_{j}^{(i)})$, whose
bounds are the ordered observed times to event.

\item \textbf{S-learner} \cite{Kunzel-etal-2018}: Instead of modeling the two
SFs independently, this method estimates a single SF, $S(t\mid \mathbf{x},A)$,
where the treatment indicator $A\in \{0,1\}$ is incorporated as an additional
feature, i.e., $\mathbf{x}^{\ast}=(\mathbf{x},A)\in \mathbb{R}^{d+1}$. This
transforms the original feature set into an augmented dataset:
\begin{equation}
\mathcal{D}^{\ast}=\{(\mathbf{x}_{1}^{\ast},y_{1},\xi_{1}),\ldots
,(\mathbf{x}_{c}^{\ast},y_{c},\xi_{c}),(\mathbf{z}_{1}^{\ast},h_{1},\delta
_{1}),\ldots,(\mathbf{z}_{t}^{\ast},h_{t},\delta_{t})\}.
\end{equation}
The CATE is then computed as:
\begin{equation}
\tau(\mathbf{x})=\sum_{j=1}^{t}(t_{j}^{(1)}-t_{j-1}^{(1)})\tilde{S}%
^{(j)}(\mathbf{x},1)-\sum_{k=1}^{c}(t_{k}^{(0)}-t_{k-1}^{(0)})\tilde{S}%
^{(k)}(\mathbf{x},0).
\end{equation}

\end{enumerate}

As a result, we have six combinations of meta-learners (T and S) and the SF
estimators (RSF, Cox, Beran) denoted as: T-RSF, S-RSF, T-Cox, S-Cox, T-Beran, S-Beran.

\subsection{Generating Synthetic Datasets}

As discussed above, we generate artificial complex feature spaces and outcomes
for our numerical experiments. All feature vectors, encompassing both controls
$\mathbf{x}$ and treatments $\mathbf{y}$, are produced using four distinct
functions: the linear function, the spiral function, the bell-shaped function,
and the circular function. These functions are selected specifically to create
complex data structures that pose challenges for many standard processing
methods. They are similar to functions proposed in \cite{Kirpichenko-etal-24}.
Each function is defined through a parameter $\zeta$ as follows:

\begin{enumerate}
\item \textbf{Linear functions:} The control and treatment effect applied to
continuous features $x^{(i)}\sim \mathcal{U}(0,1)$:
\begin{equation}
y(\mathbf{x})=2x^{(1)}+4x^{(2)}, \label{linear_func_y}%
\end{equation}%
\begin{equation}
h(\mathbf{x})=2(1-TP)\cdot x^{(1)}+4x^{(2)}+8\cdot TP\cdot x^{(3)},
\label{linear_func_h}%
\end{equation}
with $A\in \{0,1\}$ and $TP\in \{0.1,0.2,...,0.6\}$. Here $TP$ is the treatment
power, i.e., a parameter that determines how strong or effective a treatment
is, such as how many tablets affect a patient. In other words, $TP$ shows how
the number of tablets affects $h(\mathbf{x})$. It is used to study and compare
how the prediction accuracy of different models depends on treatment effect.

\item \textbf{Spiral functions:} Feature vectors of dimensionality $d=7$
positioned on Archimedean spirals are defined for even $d$ as
\begin{equation}
\mathbf{x}=(\zeta \sin(\zeta),\zeta \cos(\zeta),\ldots,\zeta \sin(\zeta \cdot
d/2),\zeta \cos(\zeta \cdot d/2)),
\end{equation}
and for odd $d$ as
\begin{equation}
\mathbf{x}=(\zeta \sin(\zeta),\zeta \cos(\zeta),\ldots,\zeta \sin(\zeta
\cdot \lceil d/2\rceil)).
\end{equation}
For all numerical experiments, values of $\zeta$ are uniformly sampled from
the interval $[0,10]$.

\item \textbf{Bell-shaped functions:} The features are constructed as a
collection of nearly non-overlapping Gaussian functions. Since $\zeta$ is
drawn from a uniform distribution, we denote $\zeta_{\min}$ and $\zeta_{\max}$
as the lower and upper bounds of this distribution. The feature vector of
dimensionality $d=6$ is then given by:
\begin{align}
\mathbf{x}  &  =(x_{1},x_{1},\ldots,x_{d}),\nonumber \\
\sigma &  =\frac{\zeta_{\max}-\zeta_{\min}}{6d},\quad \mu=\frac{\zeta_{\max
}-\zeta_{\min}}{d},\nonumber \\
x_{i}  &  =\frac{1}{\sigma \sqrt{2\pi}}\cdot \exp \left(  -\frac{(\zeta-i\cdot
\mu)^{2}}{2\sigma^{2}}\right)  ,\ i=1,\ldots,d.
\end{align}
Consequently, each feature $x_{i}$ corresponds to a distinct region within the
$\zeta$ distribution.

\item \textbf{Circular functions:} The feature space is generated using only
an even number of dimensions. Feature vectors are placed on two-dimensional
circles as follows:
\begin{align}
c_{num}  &  =\frac{d}{2},\quad c_{range}=\frac{\zeta_{\max}-\zeta_{\min}}%
{d/2},\nonumber \\
\mathbf{x}  &  =(x_{1}^{1},x_{1}^{2},x_{2}^{1},x_{2}^{2},\ldots,x_{d/2}%
^{1},x_{d/2}^{2}),\nonumber \\
x_{i}^{(1)}  &  =\sin \left(  \frac{2\pi(\zeta-(i-1)\cdot c_{range})}%
{c_{range}}\right)  \cdot \mathbf{I}_{i},\nonumber \\
x_{i}^{(2)}  &  =\cos \left(  \frac{2\pi(\zeta-(i-1)\cdot c_{range})}%
{c_{range}}\right)  \cdot \mathbf{I}_{i},\nonumber \\
\mathbf{I}_{i}  &  =\mathbf{I}\{(i-1)\cdot c_{range}\leq \zeta<i\cdot
c_{range}\},~i=1,\ldots,d/2,
\end{align}
where $d=10$, $\mathbf{I}$ denotes the indicator function. Each pair of
features $(x_{i}^{(1)},x_{i}^{(2)})$ corresponds to its own two-dimensional
circle and associated region of the $\zeta$ distribution.
\end{enumerate}

In all experiments except for the linear function dataset, feature vectors
$\mathbf{z}$ for the treatment group are generated identically to the control
vectors $\mathbf{x}$. However, the corresponding event times $y$ and $h$
differ between the groups and are sampled from Weibull distributions as
follows:
\begin{equation}
y(\zeta)=-\left(  \frac{\log(u)}{0.0005\cdot \exp(1.6\cdot \zeta)}\right)
^{1/2},
\end{equation}%
\begin{equation}
h(\zeta)=-\left(  \frac{\log(u)}{0.005\cdot \exp(0.8\cdot \zeta)}\right)
^{1/2},
\end{equation}
where $u\sim \mathcal{U}(0,1)$ is a random variable uniformly distributed on
$(0,1)$. Any values of $y$ and $h$ exceeding $2000$ are truncated to this
upper bound.

This generation scheme for $y$ and $h$ is consistent with the Cox proportional
hazards model, which justifies the inclusion of the Cox model as a baseline
alongside RSFs and the Beran estimator with Gaussian kernels in our numerical experiments.

In the linear function dataset, values $y(\mathbf{x})$ and $h(\mathbf{x})$ are
derived from the corresponding expressions (\ref{linear_func_y}) and
(\ref{linear_func_h}).

The proportion of censored observations, denoted as $p$, is fixed at $30\%$ of
all samples across all experiments. Accordingly, the censoring indicators
$\delta_{i}$ and $\xi_{i}$ are drawn from binomial distributions with
probabilities $\Pr \{ \delta_{i}=1\}=\Pr \{ \xi_{i}=1\}=0.70$ and $\Pr \{
\delta_{i}=0\}=\Pr \{ \xi_{i}=0\}=0.3$.

\subsection{Results on Benchmark Datasets}

The experiments evaluate model performance under varying censoring rates (CR),
treatment power (TP), and treatment group sizes (TR).%

\begin{figure}
[h!]
\begin{center}
\includegraphics[
height=2.1998in,
width=6.2464in
]%
{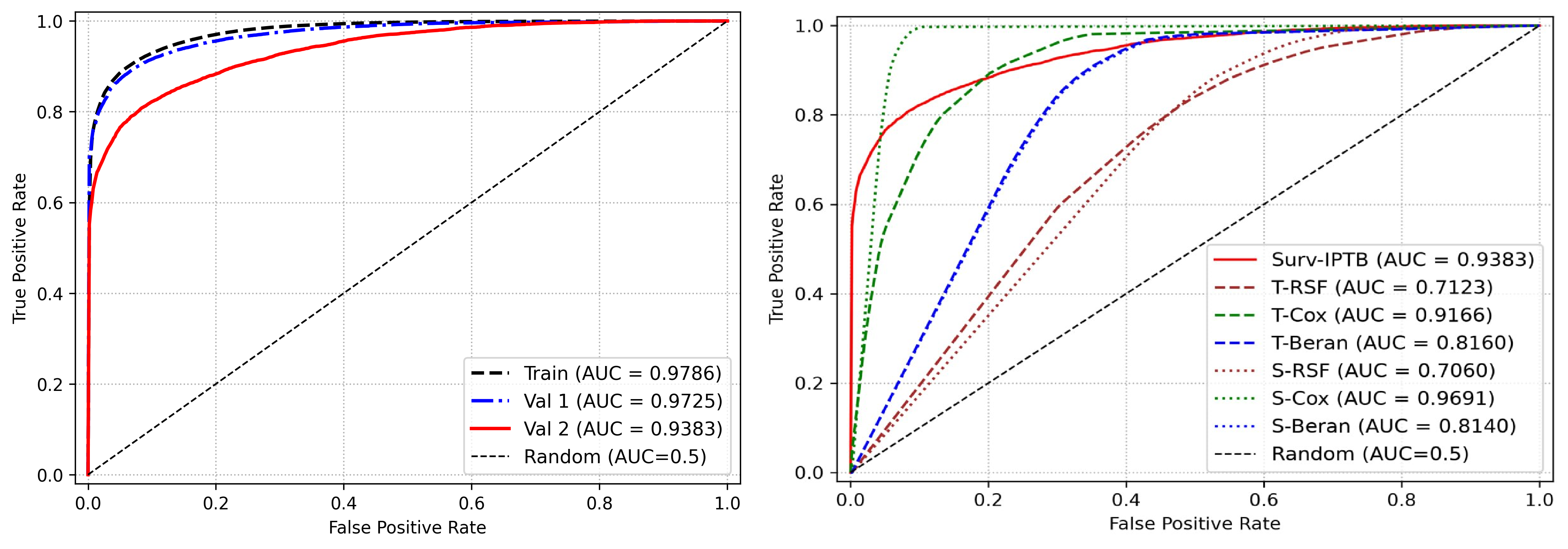}%
\caption{Left plot: The ROC curves and ROC-AUC scores obtained on the training
set and sets Val 1 and Val 2 for Surv-IPTB trained on the Linear dataset.
Right plot: Comparison of ROC curves and ROC-AUC scores for Surv-IPTB and the
meta-learners}%
\label{f:linear_surv_iptb}%
\end{center}
\end{figure}
%

\begin{table}[h!] \centering
\caption{AUC values for models trained on the linear dataset by different values of censoring rate (CR)}\resizebox{\textwidth}{!}{
\begin{tabular}
[c]{ccccccc}\hline
CR & $0.0$ & $0.1$ & $0.2$ & $0.3$ & $0.4$ & $0.5$\\ \hline
Surv-IPTB & $0.957\pm0.039$ & $0.958\pm0.019$ & $0.947\pm0.023$ &
$0.938\pm0.025$ & $0.925\pm0.031$ & $0.908\pm0.043$\\ \hline
T-RSF & $0.847\pm0.091$ & $0.792\pm0.118$ & $0.720\pm0.088$ & $0.712\pm0.095$
& $0.663\pm0.083$ & $0.642\pm0.082$\\ \hline
T-Cox & $0.924\pm0.034$ & $0.937\pm0.018$ & $0.929\pm0.028$ & $0.917\pm0.048$
& $0.918\pm0.037$ & $0.869\pm0.054$\\ \hline
T-Beran & $0.856\pm0.037$ & $0.839\pm0.036$ & $0.828\pm0.039$ & $0.816\pm
0.038$ & $0.786\pm0.049$ & $0.764\pm0.052$\\ \hline
S-RSF & $0.728\pm0.054$ & $0.717\pm0.071$ & $0.716\pm0.054$ & $0.706\pm0.056$
& $0.683\pm0.070$ & $0.648\pm0.067$\\ \hline
S-Cox & $0.976\pm0.006$ & $0.975\pm0.005$ & $0.972\pm0.005$ & $0.969\pm0.009$
& $0.962\pm0.010$ & $0.954\pm0.014$\\ \hline
S-Beran & $0.848\pm0.037$ & $0.834\pm0.037$ & $0.825\pm0.040$ & $0.814\pm
0.040$ & $0.786\pm0.048$ & $0.766\pm0.051$\\ \hline
\end{tabular}
} \label{t:Linear_CR}%
\end{table}%
%

\begin{table}[h!] \centering
\caption{AUC values for models trained on the linear dataset by different values of treatment power (TP)}
\resizebox{\textwidth}{!}{
\begin{tabular}
[c]{lcccccc}\hline
TP & $0.1$ & $0.2$ & $0.3$ & $0.4$ & $0.5$ & $0.6$\\ \hline
Surv-IPTB & $0.876\pm0.032$ & $0.911\pm0.020$ & $0.920\pm0.022$ &
$0.939\pm0.021$ & $0.942\pm0.022$ & $0.938\pm0.025$\\ \hline
T-RSF & $0.686\pm0.094$ & $0.729\pm0.099$ & $0.734\pm0.096$ & $0.726\pm0.095$
& $0.730\pm0.099$ & $0.712\pm0.095$\\ \hline
T-Cox & $0.841\pm0.062$ & $0.874\pm0.048$ & $0.894\pm0.046$ & $0.905\pm0.049$
& $0.913\pm0.047$ & $0.917\pm0.048$\\ \hline
T-Beran & $0.696\pm0.033$ & $0.755\pm0.034$ & $0.783\pm0.036$ & $0.797\pm
0.040$ & $0.808\pm0.040$ & $0.816\pm0.038$\\ \hline
S-RSF & $0.605\pm0.082$ & $0.609\pm0.066$ & $0.624\pm0.059$ & $0.667\pm0.056$
& $0.695\pm0.050$ & $0.706\pm0.056$\\ \hline
S-Cox & $0.869\pm0.025$ & $0.924\pm0.017$ & $0.943\pm0.014$ & $0.953\pm0.010$
& $0.962\pm0.010$ & $0.969\pm0.009$\\ \hline
S-Beran & $0.693\pm0.033$ & $0.753\pm0.035$ & $0.779\pm0.035$ & $0.795\pm
0.040$ & $0.807\pm0.039$ & $0.814\pm0.040$\\ \hline
\end{tabular}
} \label{t:Linear_INT}%
\end{table}%

\begin{figure}
[h!]
\begin{center}
\includegraphics[
height=2.7593in,
width=3.7754in
]%
{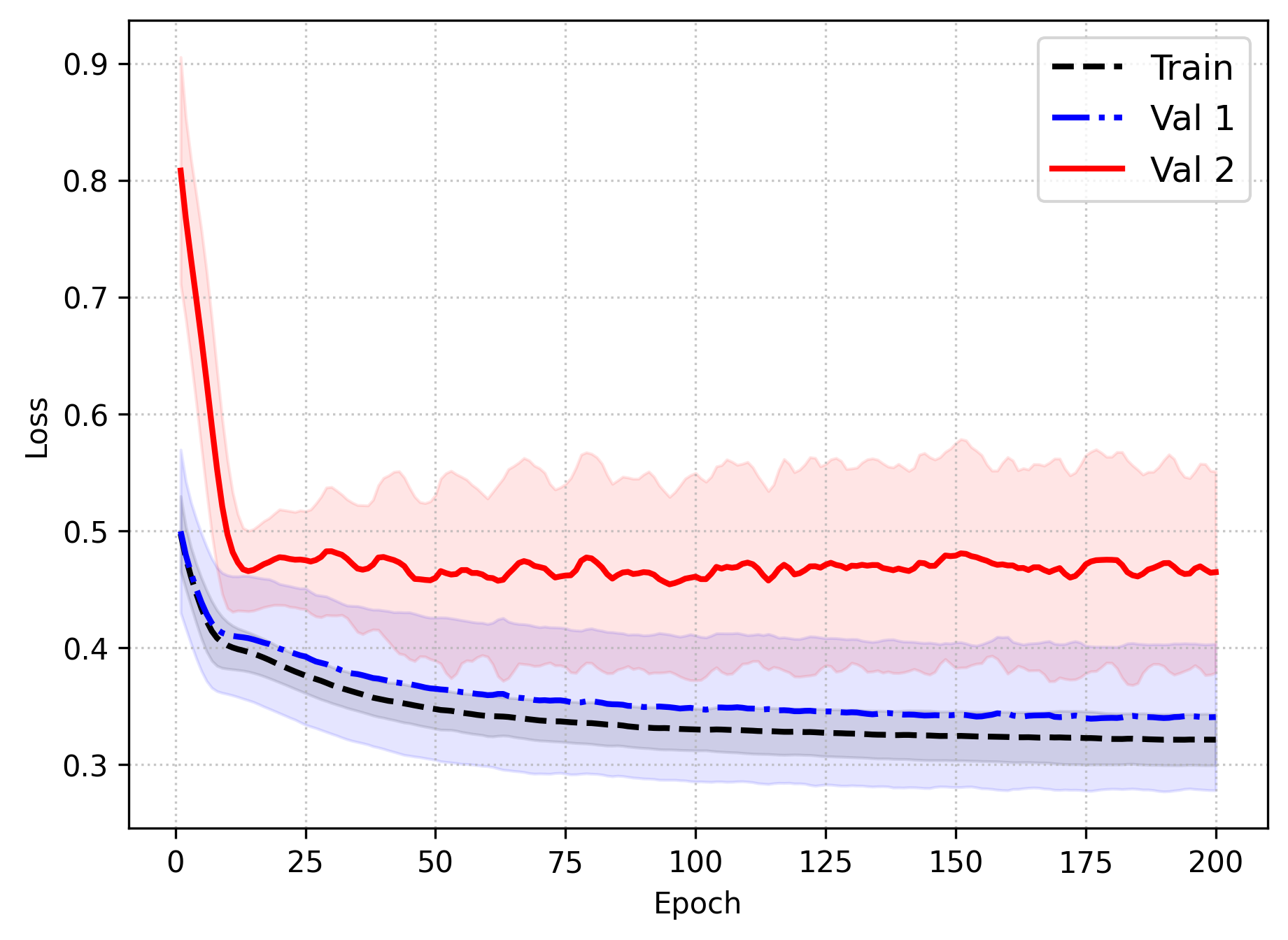}%
\caption{Training and validation loss functions for the linear dataset}%
\label{f:linear_loss}%
\end{center}
\end{figure}

The \textbf{Linear dataset} serves as a control experiment where the true
relationships are linear. Fig. \ref{f:linear_surv_iptb} illustrates the ROC
curves and ROC-AUC scores obtained on the training set and sets Val 1 and Val
2 for Surv-IPTB trained on the Linear dataset by the censoring rate 30\% (Left
plot) and the corresponding ROC curves and ROC-AUC scores for Surv-IPTB and
the meta-learners (Right plot). Fig. \ref{f:linear_loss} illustrates values of
training (Train) and two testing (Val 1 and Val 2) loss functions depending on
the epoch numbers for the linear dataset. Curves Train and Val 1 show a
consistent, gradual decline ending near $0.32$ and $0.34$, respectively,
whereas the Val 2 curve drops rapidly at the start but flattens out much
higher around $0.47$ with noticeable noise. The Val 2 loss is the highest and
has the widest confidence interval (shaded region), suggesting the model
struggles most with the second validation set.

Table \ref{t:Linear_CR} shows how the model performance depends on the
censoring rate (CR). Table \ref{t:Linear_INT} shows how the model performance
depends on the treatment power (TP) when the censoring rate is 30\%.%

On linear data, S-Cox outperfoms all models achieving AUCs of $0.976$ at
CR=0.0 and maintaining $0.954$ at CR=0.5, as the S-learner's pooled data
approach combined with a correctly-specified Cox model proves maximally
efficient. Surv-IPTB performs very well with AUC $\sim0.94$ across all CRs but
is consistently outperformed by S-Cox, confirming that simpler,
correctly-specified models beat complex ones on simple data. T-Cox performs
well initially but shows a sharper decline with censoring, dropping from
$0.924$ to $0.869$, due to reduced sample sizes from splitting data into two
groups. RSF-based models (T-RSF, S-RSF) perform worst on linear data with AUCs
in the $0.64$--$0.72$ range, demonstrating that tree-based methods are
data-hungry and suffer high variance on low-dimensional, simple structures.
Regarding treatment power, as TP increases, all models improve, but S-Cox
shows the most dramatic gain from $0.869$ to $0.969$, perfectly exploiting the
strengthening linear signal, while Surv-IPTB improves more modestly from
$0.876$ to $0.938$, indicating it is less sensitive to treatment effect strength.%

\begin{table}[h!] \centering
\caption{AUC values for models trained on the bell dataset by different values of censoring rate (CR)}
\resizebox{\textwidth}{!}{
\begin{tabular}
[c]{lcccccc}\hline
CR & $0.0$ & $0.1$ & $0.2$ & $0.3$ & $0.4$ & $0.5$\\ \hline
Surv-IPTB & $0.940\pm0.022$ & $0.901\pm0.113$ & $0.940\pm0.025$ &
$0.920\pm0.086$ & $0.932\pm0.027$ & $0.927\pm0.035$\\ \hline
T-RSF & $0.906\pm0.024$ & $0.843\pm0.036$ & $0.774\pm0.058$ & $0.721\pm0.065$
& $0.692\pm0.044$ & $0.648\pm0.052$\\ \hline
T-Cox & $0.683\pm0.031$ & $0.678\pm0.026$ & $0.660\pm0.019$ & $0.646\pm0.019$
& $0.638\pm0.026$ & $0.623\pm0.025$\\ \hline
T-Beran & $0.693\pm0.022$ & $0.687\pm0.022$ & $0.678\pm0.027$ & $0.668\pm
0.028$ & $0.656\pm0.025$ & $0.634\pm0.030$\\ \hline
S-RSF & $0.901\pm0.031$ & $0.836\pm0.047$ & $0.761\pm0.061$ & $0.720\pm0.079$
& $0.705\pm0.064$ & $0.666\pm0.065$\\ \hline
S-Cox & $0.730\pm0.037$ & $0.727\pm0.042$ & $0.731\pm0.044$ & $0.724\pm0.043$
& $0.730\pm0.039$ & $0.721\pm0.041$\\ \hline
S-Beran & $0.696\pm0.020$ & $0.689\pm0.025$ & $0.668\pm0.026$ & $0.660\pm
0.031$ & $0.649\pm0.030$ & $0.617\pm0.027$\\ \hline
\end{tabular}
} \label{t:Bell_CR}%
\end{table}%

\begin{figure}
[h!]
\begin{center}
\includegraphics[
height=2.1548in,
width=5.86in
]%
{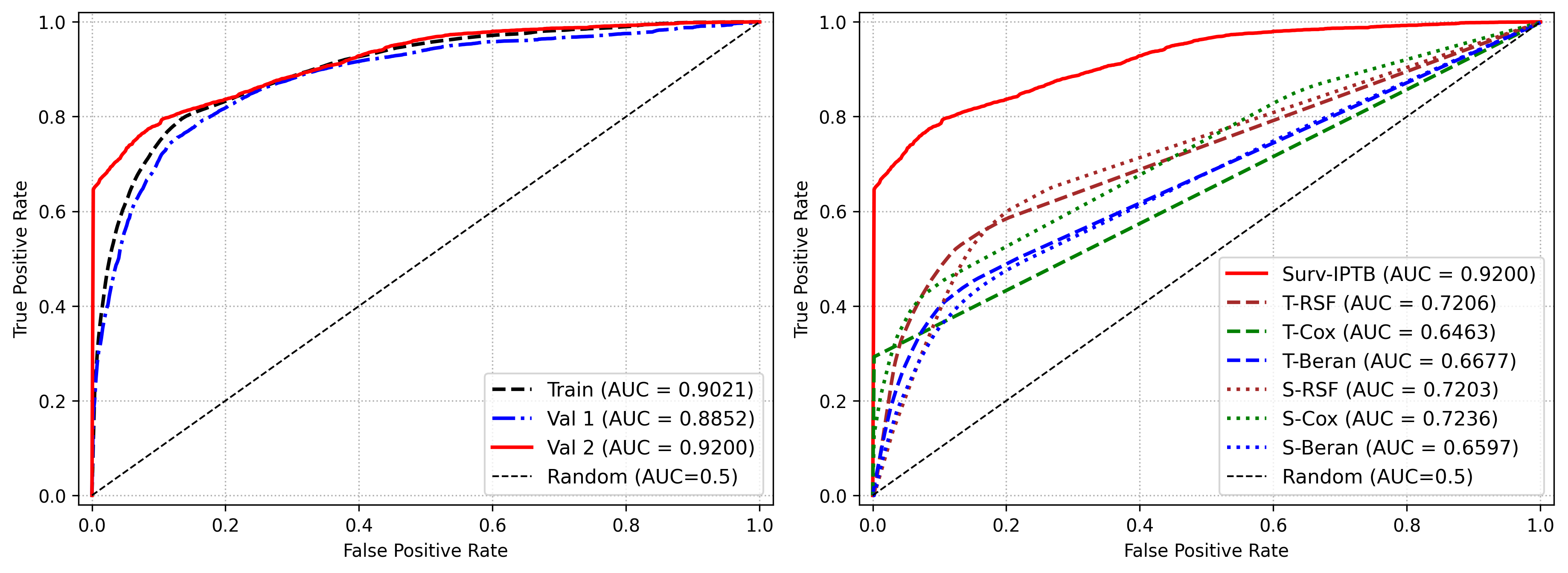}%
\caption{Left plot: The ROC curves and ROC-AUC scores obtained on the training
set and sets Val 1 and Val 2 for Surv-IPTB trained on the Bell-shaped dataset.
Right plot: Comparison of ROC curves and ROC-AUC scores for Surv-IPTB and the
meta-learners}%
\label{f:bell_surv_iptb}%
\end{center}
\end{figure}

The next dataset is \textbf{Bell-shaped}. Fig. \ref{f:bell_surv_iptb}
illustrates the ROC curves obtained on the training set and sets Val 1 and Val
2 for Surv-IPTB trained on the Bell-shaped dataset by the censoring rate 30\%
(Left plot) and the corresponding ROC curves for Surv-IPTB and the
meta-learners (Right plot).%

\begin{figure}
[h!]
\begin{center}
\includegraphics[
height=2.8971in,
width=3.9619in
]%
{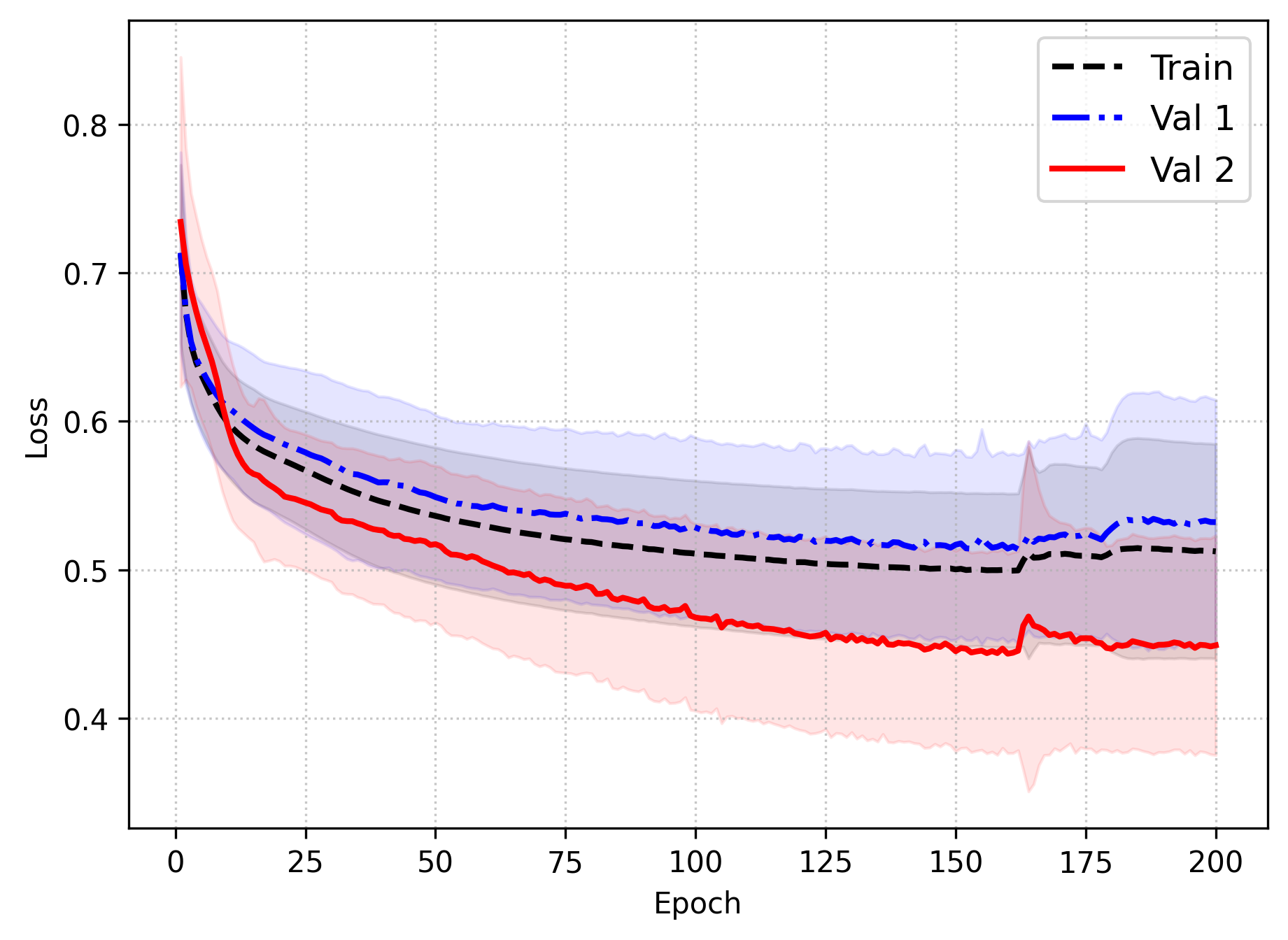}%
\caption{Training and validation loss functions for the Bell-shaped dataset}%
\label{f:bell_loss}%
\end{center}
\end{figure}

Fig. \ref{f:bell_loss} illustrates values of training (Train) and two testing
(Val 1 and Val 2) loss functions depending on the epoch numbers for the
Bell-shaped dataset. The graph plots loss show a sharp initial decline for all
curves during the first 50 epochs followed by a gradual stabilization, with
shaded areas indicating variance or confidence intervals.%

Table \ref{t:Bell_CR} shows how the model performance depends on the censoring
rate. RSF models become competitive here, with T-RSF and S-RSF starting at
$0.906$--$0.901$ and declining more gracefully, as the localized feature
structure is ideally suited for tree-based partitioning. Surv-IPTB
nevertheless remains the best, consistently achieving AUCs betwenn $0.93$ and
$0.94$ across all CR levels and demonstrating superior generalization even
when RSFs improve. Cox and Beran models perform poorly with AUCs around
$0.6-0.7$ across all variants.%

\begin{table}[h!] \centering
\caption{AUC values for models trained on the circular dataset by different values of censoring rate (CR). Surv-IPTB is implemented with trainable values of $\pi ^{(r,s)}$ and with approximate using these values obtained by means of the Kaplan-Meier estimator}
\resizebox{\textwidth}{!}{
\begin{tabular}
[c]{lcccccc}\hline
CR & $0.0$ & $0.1$ & $0.2$ & $0.3$ & $0.4$ & $0.5$\\ \hline
Surv-IPTB & $0.983\pm0.010$ & $0.979\pm0.015$ & $0.983\pm0.011$ &
$0.976\pm0.015$ & $0.971\pm0.015$ & $0.968\pm0.016$\\ \hline
Surv-IPTB-KM & $0.983\pm0.010$ & $0.981\pm0.011$ & $0.981\pm0.013$ &
$0.977\pm0.013$ & $0.970\pm0.019$ & $0.966\pm0.019$\\ \hline
T-RSF & $0.909\pm0.044$ & $0.832\pm0.054$ & $0.782\pm0.051$ & $0.718\pm0.043$
& $0.681\pm0.053$ & $0.637\pm0.071$\\ \hline
T-Cox & $0.539\pm0.051$ & $0.539\pm0.053$ & $0.557\pm0.051$ & $0.558\pm0.047$
& $0.552\pm0.055$ & $0.565\pm0.035$\\ \hline
T-Beran & $0.865\pm0.042$ & $0.836\pm0.042$ & $0.803\pm0.042$ & $0.773\pm
0.048$ & $0.756\pm0.051$ & $0.715\pm0.044$\\ \hline
S-RSF & $0.924\pm0.039$ & $0.857\pm0.049$ & $0.775\pm0.078$ & $0.737\pm0.080$
& $0.705\pm0.094$ & $0.674\pm0.087$\\ \hline
S-Cox & $0.540\pm0.051$ & $0.535\pm0.040$ & $0.531\pm0.036$ & $0.529\pm0.060$
& $0.545\pm0.077$ & $0.549\pm0.080$\\ \hline
S-Beran & $0.871\pm0.041$ & $0.835\pm0.042$ & $0.791\pm0.045$ & $0.749\pm
0.057$ & $0.722\pm0.051$ & $0.675\pm0.043$\\ \hline
\end{tabular}
} \label{t:Circular_CR_no_Pi}%
\end{table}%
%

\begin{table}[h!] \centering
\caption{AUC values for models trained on the Circular dataset by different values of treatment group size (TR)}
\resizebox{\textwidth}{!}{
\begin{tabular}
[c]{lccccccc}\hline
TR & $200$ & $175$ & $150$ & $125$ & $100$ & $75$ & $50$\\ \hline
Surv-IPTB & $0.976\pm0.015$ & $0.981\pm0.011$ & $0.978\pm0.016$ &
$0.971\pm0.020$ & $0.973\pm0.019$ & $0.966\pm0.019$ & $0.909\pm0.154$\\ \hline
T-RSF & $0.823\pm0.062$ & $0.752\pm0.065$ & $0.747\pm0.054$ & $0.731\pm0.064$
& $0.718\pm0.043$ & $0.718\pm0.062$ & $0.709\pm0.059$\\ \hline
T-Cox & $0.558\pm0.047$ & $0.571\pm0.052$ & $0.549\pm0.049$ & $0.552\pm0.043$
& $0.564\pm0.050$ & $0.541\pm0.062$ & $0.540\pm0.045$\\ \hline
T-Beran & $0.773\pm0.048$ & $0.774\pm0.042$ & $0.771\pm0.052$ & $0.747\pm
0.050$ & $0.762\pm0.039$ & $0.759\pm0.048$ & $0.767\pm0.068$\\ \hline
S-RSF & $0.769\pm0.077$ & $0.755\pm0.055$ & $0.747\pm0.072$ & $0.743\pm0.069$
& $0.743\pm0.119$ & $0.737\pm0.080$ & $0.735\pm0.066$\\ \hline
S-Cox & $0.529\pm0.060$ & $0.547\pm0.064$ & $0.537\pm0.060$ & $0.538\pm0.065$
& $0.501\pm0.056$ & $0.511\pm0.062$ & $0.512\pm0.084$\\ \hline
S-Beran & $0.749\pm0.057$ & $0.750\pm0.051$ & $0.730\pm0.063$ & $0.723\pm
0.066$ & $0.715\pm0.047$ & $0.696\pm0.052$ & $0.649\pm0.079$\\ \hline
\end{tabular}
} \label{t:Circular_TR}%
\end{table}%

Let us consider the \textbf{Circular dataset}. Results of the model comparison
are shown in Tables \ref{t:Circular_CR_no_Pi} and \ref{t:Circular_TR}. They
illustrate how the model performance depend on the censoring rate (Table
\ref{t:Circular_CR_no_Pi}) and on the treatment group size (Table
\ref{t:Circular_TR}).

The circular structure is very difficult for linear models, yet tree-based
methods perform well on this data. Surv-IPTB dominates this setting,
maintaining AUCs above $0.96$ even at large censoring rate (CR=0.5) and
showing exceptional robustness. We compare two implementations of Surv-IPTB.
The first implemetation, Surv-IPTB, supposes that $\pi^{(r,s)}$,
$(r,s)\in \mathcal{T}^{-}\cup \mathcal{C}^{+}$, is trained in accordance with
the given loss functions jointly with parameters of the attention
$\mathbf{W}_{Q}$ and $\mathbf{W}_{K}$. The second implementation,
Surv-IPTB-KM, does not train $\pi^{(r,s)}$ and uses values derived from the
Kaplan-Meier estimator. These implementations are provided to study how
Surv-IPTB-KM approximate the full training implementation and can be used this
approximation instead of Surv-IPTB. Table \ref{t:Circular_CR_no_Pi} shows the
difference of the implementations by different values of censoring rate. One
can see from Table \ref{t:Circular_CR_no_Pi} that Surv-IPTB-KM performs almost identically.

In stark contrast, RSF models collapse dramatically, starting at
$0.909$--$0.924$ at CR=0.0 and plummeting to $0.64$--$0.67$ at CR=0.5, as the
forests overfit the circular pattern and become unstable with missing data.
Cox models prove essentially useless with AUCs around $0.53$--$0.56$,
representing near-random performance that confirms any non-linear structure
destroys Cox regression. Regarding treatment group size, Surv-IPTB remains
stable until TR drops to $50$ (AUC $0.909$ with high variance $\pm0.154$),
while other models show no sensitivity to TR, indicating their limitations are
structural rather than data-driven.%

\begin{table}[h!] \centering
\caption{AUC values for models trained on the Spiral dataset by different values of censoring rate (CR). Surv-IPTB is implemented with trainable values of $\pi ^{(r,s)}$ and with approximate using these values obtained by means of the Kaplan-Meier estimator}
\resizebox{\textwidth}{!}{
\begin{tabular}
[c]{lcccccc}\hline
CR & $0.0$ & $0.1$ & $0.2$ & $0.3$ & $0.4$ & $0.5$ \\ \hline
Surv-IPTB & $0.937\pm0.037$ & $0.901\pm0.022$ & $0.971\pm0.040$ &
$0.963\pm0.053$ & $0.894\pm0.142$ & $0.956\pm0.056$\\ \hline
Surv-IPTB-KM & $0.937\pm0.037$ & $0.901\pm0.022$ & $0.949\pm0.057$ &
$0.979\pm0.021$ & $0.921\pm0.076$ & $0.928\pm0.081$\\ \hline
T-RSF & $0.931\pm0.034$ & $0.854\pm0.039$ & $0.796\pm0.065$ & $0.727\pm0.047$
& $0.705\pm0.038$ & $0.663\pm0.041$\\ \hline
T-Cox & $0.652\pm0.053$ & $0.653\pm0.049$ & $0.641\pm0.042$ & $0.636\pm0.054$
& $0.634\pm0.041$ & $0.614\pm0.035$\\ \hline
T-Beran & $0.920\pm0.045$ & $0.900\pm0.045$ & $0.873\pm0.045$ & $0.854\pm
0.043$ & $0.826\pm0.054$ & $0.793\pm0.050$\\ \hline
S-RSF & $0.935\pm0.039$ & $0.879\pm0.044$ & $0.803\pm0.058$ & $0.761\pm0.067$
& $0.757\pm0.055$ & $0.728\pm0.062$\\ \hline
S-Cox & $0.552\pm0.129$ & $0.575\pm0.101$ & $0.596\pm0.102$ & $0.581\pm0.091$
& $0.593\pm0.096$ & $0.608\pm0.102$\\ \hline
S-Beran & $0.926\pm0.044$ & $0.906\pm0.042$ & $0.878\pm0.042$ & $0.857\pm
0.041$ & $0.830\pm0.053$ & $0.798\pm0.048$\\ \hline
\end{tabular}
} 
\label{t:Spiral_CR}%
\end{table}%
%

Let us study the \textbf{Spiral dataset} now. The intertwined spiral structure
is even more challenging than the circle. The corresponding results are shown
in Table \ref{t:Spiral_CR}. It can be seen from the
table that Surv-IPTB maintains a clear lead with AUCs consistently above
$0.9$ and demonstrates robustness across censoring levels. RSF models again
suffer a sharp performance drop from $0.93$ at zero censoring rate (CR=0.0) to
$0.66$--$0.72$ at CR=0.5, confirming overfitting and instability under
censoring. Interestingly, Beran models perform surprisingly well with AUCs
from $0.92$ down to $0.79$, better than on the Circle dataset, as the Gaussian
kernel smoother handles the spiral's continuous gradient more effectively than
the circle's abrupt periodicity. Cox models fail completely with AUCs stuck at
$0.55$--$0.65$, reinforcing that non-linear data destroys linear model
inference. Table \ref{t:Spiral_CR} also shows the difference of the
implementations (models Surv-IPTB and Surv-IPTB-KM) by different values of the
censoring rate.%

\begin{figure}
[h!]
\begin{center}
\includegraphics[
height=2.2368in,
width=6.0951in
]%
{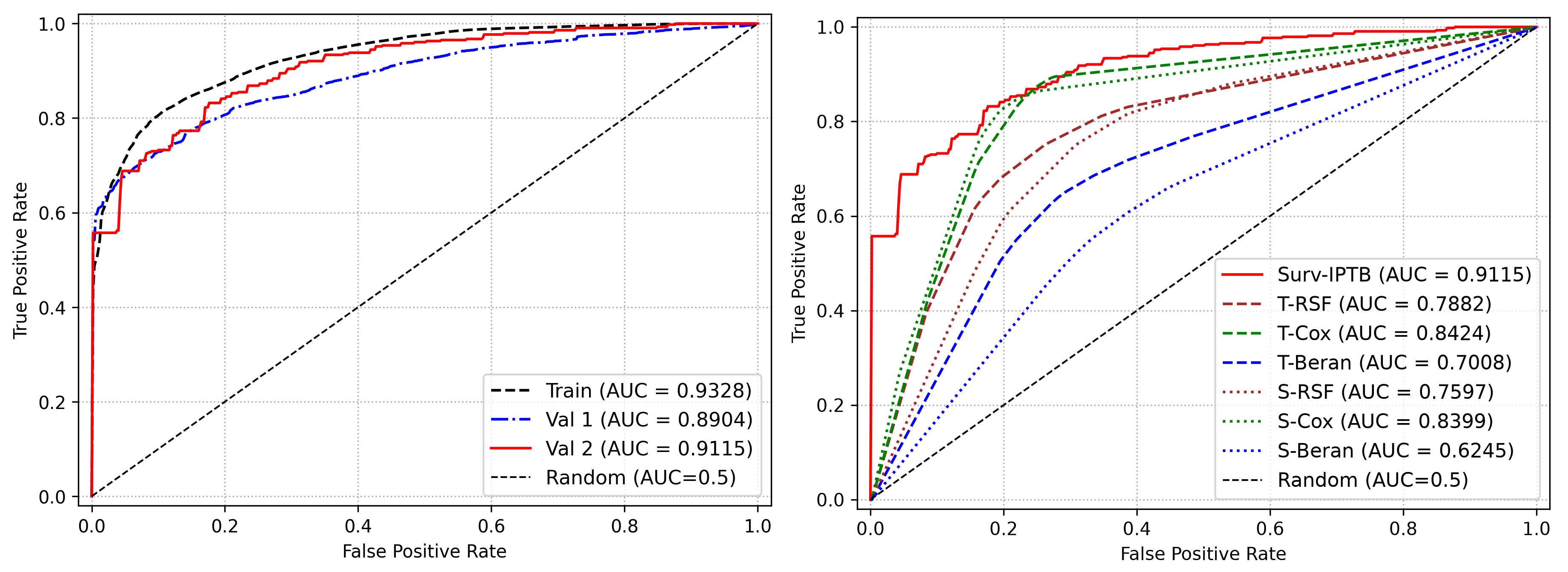}%
\caption{Left plot: The ROC curves and ROC-AUC scores obtained on the training
set and sets Val 1 and Val 2 for Surv-IPTB trained on the IHDP dataset. Right
plot: Comparison of ROC curves and ROC-AUC scores for Surv-IPTB and the
meta-learners}%
\label{f:ihdp_surv_iptb}%
\end{center}
\end{figure}

We also evaluate the performance of Surv-IPTB on the widely used Infant
\textbf{Health and Development Program (IHDP) dataset} \cite{Hill-2011}, a
semi-synthetic benchmark based on real covariates for estimating heterogeneous treatment effects (HTE). The IHDP-100 dataset comprises 747 subjects with 25 covariates: 6 continuous and
19 binary. The target variables are considered as times to events. Censored observations  are drawn from binomial distributions with
probabilities $\Pr \{ \delta_{i}=0\}=\Pr \{ \xi_{i}=0\}=0.3$.
The data is publicly available at
{\url{https://github.com/vdorie/npci}}.

Fig. \ref{f:ihdp_surv_iptb} illustrates the ROC curves obtained on the
training, Val 1, and Val 2 sets for Surv-IPTB trained on the IHDP-100 dataset
by the censoring rate 30\% (Left plot) and the corresponding ROC curves for
Surv-IPTB and the meta-learners (Right plot).

Table \ref{t:IHDP_results} illustrates how the performance of models trained
on the IHDP-100 dataset depend on the censoring rate. It demonstrates that
Surv-IPTB consistently outperforms all baseline methods across all censoring
rates, maintaining high AUC values between $0.898$ and $0.916$ with remarkably
low variance, whereas the T-RSF and T-Cox models show near-random performance
at zero censoring and fail to match Surv-IPTB's robustness as censoring
increases. Notably, while most models exhibit modest performance degradation
with higher censoring rates, the kernel-based Beran estimators (both T-Beran
and S-Beran) deliver the poorest results, suggesting that the IHDP-100 data
structure is too complex for simple local smoothing approaches to capture effectively.%

\begin{table}[h!] \centering
\caption{AUC values for models trained on the the IHDP-100 dataset by different values of censoring rate (CR)}
\resizebox{\textwidth}{!}{
\begin{tabular}{lcccccc}
\hline
CR & 0.0 & 0.1 & 0.2 & 0.3 & 0.4 & 0.5 \\ \hline
Surv-IPTB & $0.916 \pm 0.036$ & $0.914 \pm 0.037$ & $0.915 \pm 0.035$ & $0.912 \pm 0.040$ & $0.908 \pm 0.045$ & $0.898 \pm 0.057$ \\ \hline
T-RSF & $0.500 \pm 0.000$ & $0.776 \pm 0.058$ & $0.808 \pm 0.040$ & $0.788 \pm 0.049$ & $0.796 \pm 0.037$ & $0.770 \pm 0.077$ \\ \hline
T-Cox & $0.500 \pm 0.000$ & $0.833 \pm 0.043$ & $0.846 \pm 0.036$ & $0.842 \pm 0.041$ & $0.846 \pm 0.036$ & $0.841 \pm 0.040$ \\ \hline
T-Beran & $0.714 \pm 0.065$ & $0.711 \pm 0.065$ & $0.725 \pm 0.076$ & $0.701 \pm 0.071$ & $0.697 \pm 0.066$ & $0.690 \pm 0.058$ \\ \hline
S-RSF & $0.768 \pm 0.054$ & $0.780 \pm 0.058$ & $0.781 \pm 0.056$ & $0.760 \pm 0.050$ & $0.775 \pm 0.060$ & $0.760 \pm 0.045$ \\ \hline
S-Cox & $0.843 \pm 0.046$ & $0.840 \pm 0.044$ & $0.837 \pm 0.048$ & $0.840 \pm 0.046$ & $0.848 \pm 0.039$ & $0.828 \pm 0.047$ \\ \hline
S-Beran & $0.687 \pm 0.071$ & $0.665 \pm 0.075$ & $0.677 \pm 0.072$ & $0.624 \pm 0.043$ & $0.613 \pm 0.064$ & $0.612 \pm 0.078$ \\ \hline
\end{tabular}
} \label{t:IHDP_results}%
\end{table}%

Table~\ref{t:IHDP_results} demonstrates that Surv-IPTB consistently
outperforms all baseline methods across all censoring rates, maintaining high
AUC values between 0.898 and 0.916 with remarkably low variance, whereas the
T-RSF and T-Cox models show near-random performance at zero censoring and fail
to match Surv-IPTB's robustness as censoring increases.

Finally, we provide Table \ref{tab:best_loss} which shows the best loss values
for Surv-IPTB computed based on the Train, Val1, and Val2 subsets under
condition CR=0.3. It reveals that the linear dataset achieves the lowest
training and validation losses, indicating that the model fits this simple
structure most effectively, while the spiral dataset exhibits the highest
losses with substantial variance across all subsets, reflecting the difficulty
of capturing its complex intertwined patterns. The circular dataset presents
an interesting anomaly where the Val2 loss ($0.312$) is notably lower than
both Train ($0.379$) and Val1 ($0.405$), suggesting that the Val2 subset is
constructed only from the pairs of uncensored observations. Across all
datasets except circular, the losses on Val1 and Val2 remain close to the
training loss, indicating minimal overfitting and stable generalization performance.

\begin{table}[h!]
\caption{Best loss values for Surv-IPTB on the Train, Val1, and Val2 subsets
for different datasets}%
\label{tab:best_loss}
\centering
\begin{tabular}
[c]{lccc}\hline
Dataset & Train & Val1 & Val2\\ \hline
Linear & $0.318 \pm0.020$ & $0.334 \pm0.060$ & $0.397 \pm0.046$\\ \hline
Bell-shaped & $0.503 \pm0.059$ & $0.510 \pm0.061$ & $0.436 \pm0.069$\\ \hline
Spiral & $0.564 \pm0.164$ & $0.567 \pm0.169$ & $0.554 \pm0.209$\\ \hline
Circular & $0.379 \pm0.016$ & $0.405 \pm0.034$ & $0.312 \pm0.034$\\ \hline
IHDP-100 & $0.519 \pm0.026$ & $0.527 \pm0.043$ & $0.488 \pm0.039$\\ \hline
\end{tabular}
\end{table}

Surv-IPTB emerges as the most robust and general-purpose model. While not best
on linear data where S-Cox wins, it is the only model that maintains excellent
performance across all complex scenarios. The method proves highly stable
against censoring and sample imbalance, making it the recommended choice for
real-world data. All meta-models remain less robust than Surv-IPTB due to
kernel and bandwidth sensitivity. Finally, censoring acts as the great
equalizer: all models degrade with censoring, but Surv-IPTB degrades least, as
its deep-learning architecture proves more sample-efficient and better at
learning from incomplete data.

\section{Conclusion}

This work introduced a novel attention-based framework for estimating the
Individual Probability of Treatment Benefit (IPTB) in survival analysis
reformulating the problem as binary classification with pairwise patient
comparisons. The framework provides principled handling of right-censored
observations through imprecise probability representations, using trainable
parameters $\pi^{(r,s)}$ for uncertain treatment effect statuses that are
learned jointly with attention weights. Extensive experiments across synthetic
and real datasets (linear, bell-shaped, circular, spiral, IHDP) demonstrate
that Surv-IPTB consistently outperforms meta-learner baselines, particularly
in challenging nonlinear scenarios where conventional methods exhibit
substantial degradation. The proposed model maintains exceptional stability
under increasing censoring rates, with performance decline significantly
slower than RSF-based meta-learners. Surv-IPTB also demonstrates robust
performance even with reduced treatment group sizes, maintaining stability
until group sizes drop to 50 patients, suggesting efficient learning from
limited clinical data.

The attention mechanism provides natural interpretability through its learned
weights, indicating which treatment-control pairs are most influential for
specific predictions. Comparison of trainable versus Kaplan-Meier-derived
$\pi$ parameters revealed nearly identical performance, suggesting the simpler
approximation may suffice in practice.

This approach advances personalized medicine by enabling clinicians to
estimate individual treatment benefit probabilities supporting shared
decision-making and more efficient use of censored survival data.

Despite its strengths, our framework has several limitations that suggest
promising directions for future work:

The current framework assumes independent censoring mechanisms across
treatment and control groups. In practice, censoring may depend on covariates
(informative censoring) or differ systematically between arms. Extending the
framework to handle covariate-dependent censoring through weighting or
propensity-based adjustment would enhance its real-world applicability.

The pairwise comparison nature of our approach scales quadratically with
sample size, potentially limiting applicability to large datasets. While this
is manageable for moderate-sized clinical datasets (hundreds to thousands of
patients), scaling to biobank-scale data (tens of thousands or more) would
require approximations, such as random subsampling of pairs or
locality-sensitive hashing to restrict attention to informative comparisons.

While we have focused on single-head attention, multi-head attention and
transformer architectures could capture more complex relationships between
covariates and treatment effects, potentially further improving performance on
high-dimensional data.

\bibliographystyle{unsrt}
\bibliography{Attention,Boosting,Classif_bib,Explain,Imprbib,Interv_NN,IntervalClass,MYBIB,MYUSE,Surv_Attent,Survival_analysis,Transformer,Treatment}

\end{document}